\documentclass{article} 
\usepackage{iclr2020_conference,times}

\usepackage[utf8]{inputenc} 
\usepackage[T1]{fontenc}    

\usepackage{amsmath,amsthm,amsfonts}
\usepackage{amssymb}
\usepackage{mathrsfs}
\usepackage{mathtools}
\usepackage{stmaryrd} 
\usepackage{colonequals}
\usepackage{xcolor}
\usepackage{multirow}
\usepackage{multicol}
\usepackage{pifont}
\newcommand{\cmark}{\ding{51}}%
\newcommand{\xmark}{\ding{55}}%
\usepackage{hyperref}       
\usepackage{url}            
\usepackage{booktabs}       
\usepackage[inline]{enumitem}

\usepackage{nicefrac}       
\usepackage{microtype}      

\hypersetup{
    urlcolor  = NavyBlue,  
    linkcolor = cyan,      
    citecolor = purple,    
    filecolor = magenta    
}

\usepackage{amsmath,amsfonts,bm}

\DeclareMathAlphabet{\mathsfit}{\encodingdefault}{\sfdefault}{m}{sl}
\SetMathAlphabet{\mathsfit}{bold}{\encodingdefault}{\sfdefault}{bx}{n}

\DeclareMathOperator*{\E}{\mathbb{E}}

\DeclarePairedDelimiter\roundbracket{(}{)}
\DeclarePairedDelimiter\squarebracket{[}{]}
\DeclarePairedDelimiter\curlybracket{\{}{\}}
\makeatletter
\def\rbr{\@ifnextchar[{\roundbracket}{\roundbracket*}}
\def\sbr{\@ifnextchar[{\squarebracket}{\squarebracket*}}
\def\cbr{\@ifnextchar[{\curlybracket}{\curlybracket*}}
\makeatother

\title{Bridging the domain gap in cross-lingual \\ document classification}

\author{Guokun Lai\thanks{Work completed at Facebook AI} $^1$, Barlas Oguz$^2$, Yiming Yang$^1$, Veselin Stoyanov$^2$   \\
	Carnegie Mellon University$^1$, Facebook AI$^2$  \\
	\texttt{\{guokun,yiming\}@cs.cmu.edu}$^1$, \texttt{\{barlaso,ves\}@fb.com}$^2$ \\
}
\iclrfinalcopy
\begin{document}

\maketitle

\begin{abstract}
The scarcity of labeled training data often prohibits the internationalization of NLP models to multiple languages.  Recent developments in cross-lingual understanding (XLU) has made progress in this area, trying to bridge the language barrier using language universal representations.  However, even if the language problem was resolved,  models trained in one language would not transfer to another language perfectly due to the natural domain drift across languages and cultures.  We consider the setting of semi-supervised cross-lingual understanding, where labeled data is available in a source language (English), but only unlabeled data is available in the target language.  We combine state-of-the-art cross-lingual methods with recently proposed methods for weakly supervised learning such as unsupervised pre-training and unsupervised data augmentation to simultaneously close both the language gap and the domain gap in XLU.  We show that addressing the domain gap is crucial.  We improve over strong baselines and achieve a new state-of-the-art for cross-lingual document classification.
\end{abstract}

\section{Introduction}
\label{sec:intro}


Recent advances in Natural Language Processing have enabled us to train high-accuracy systems for many language tasks. 
However, training an accurate system still requires a large amount of training data. 
It is inefficient to collect data for a new task and it is virtually impossible to annotate a separate data set for each language. 
To go beyond English and a few popular languages, we need methods that can learn from data in one language and apply it to others.  

Cross-Lingual Understanding (XLU) has emerged as a field concerned with learning models on data in one language and applying it to others. 
Much of the work in XLU focuses on the zero-shot setting, which assumes that labeled data is available in one source language (usually English) and not in any of the target languages in which the model is evaluated. 
The labeled data can be used to train a high quality model in the source language.  
One then relies on general domain parallel corpora and monolingual corpora to learn to `transfer' from the source language to the target language.  
Transfer methods can explicitly rely on machine translation models built from such parallel corpora.  
Alternatively, one can use such corpora to learn language universal representations to produce features to train a model in one language, which one can directly apply to other languages.  
Such representations can be in the form of cross-lingual word embeddings, contextual word embeddings, or sentence embeddings (\cite{ruder2017survey, lample2019cross, schwenk2017learning}).  
Using such techniques, recent work has demonstrated reasonable zero-shot performance for cross-lingual document classification (\cite{schwenk2018corpus}) and natural language inference (\cite{conneau2018xnli}). 

What we have so far described is a simplified view of XLU, which focuses solely on the problem of aligning languages.  
This view assumes that, if we had access to a perfect translation system, and translated our source training data into the target language, the resulting model would perform as well as if we had collected a similarly sized labeled dataset directly in our target language. 
Existing work in XLU to date also works under this assumption.  
However, in real world applications, we must also bridge the domain gap across different languages, as well as the language gap.  
No task is ever identical in two languages, even if we group them under the same label, e.g. `news document classification' or `product reviews'.  
A Chinese customer might express sentiment differently than his American counterpart.  
Or French news might simply cover different topics than English news.  
As a result, any approach which ignores this domain drift will fall short of native in-language performance in real world XLU.

In this paper, we propose to jointly tackle both language and domain transfer.  
We consider the semi-supervised XLU setting, where in addition to labeled data in a source language, we have access to unlabeled data in the target language.  
Using this unlabeled data, we combine the aforementioned cross-lingual methods with recently proposed unsupervised domain adaptation and weak supervision techniques on the task of cross-lingual document classification.  
In particular, we focus on two approaches for domain adaptation.  The first method is based on masked language model (MLM) pre-training (as in \cite{devlin2018bert}) using unlabeled target language corpora.  
Such methods have been shown to improve over general purpose pre-trained models such as BERT in the weakly supervised setting (\cite{lee2019biobert, han2019unsupervised}).  
The second method is unsupervised data augmentation (UDA) (\cite{xie2019unsupervised}), where synthetic paraphrases are generated from the unlabeled corpus, and the model is trained on a label consistency loss.

While both of these techniques were proposed previously, in both cases it is non-trivial to extend them to the cross-lingual setting.  
For instance when performing data augmentation, one could generate paraphrases in either the source or the target language or both. 
We experiment with various approaches and provide guidelines with ablation studies. 
Furthermore, we find that the value of additional labeled data in the source language is limited due to the train-test discrepancy of XLU tasks. 
We propose to alleviate this issue by using self-training technique to do the domain adaptation from the source language into the target language. 
By combining these methods, we are able to reduce error rates by an average 44\% over a strong XLM baseline, setting a new state-of-the-art for cross-lingual document classification.

\section{Related Work}

\subsection{Cross-lingual understanding}
Cross-lingual document classification was first introduced in \citep{bel2003cross}.
The subsequent work \citep{prettenhofer2010cross} proposes the cross-lingual sentiment classification datasets, and \citep{lewis2004rcv1, klementiev2012inducing, schwenk2018corpus} have extended this to the news domain.  Cross-lingual understanding has also been applied to other NLP tasks, with datasets available in dependency parsing \citep{nivre2016universal}, natural language inference (XNLI) \citep{conneau2018xnli} and question answering (\cite{liu2019xqa}).

Cross-lingual methods gained popularity with the advent of cross-lingual word embeddings (\cite{mikolov2013exploiting}).  Since then, many methods have been proposed to better align the word embedding spaces of different languages (see \cite{ruder2017survey} for a survey).  Recently, more sophisticated extensions have been proposed based on seq2seq training of cross-lingual sentence embeddings (\cite{schwenk2017learning, artetxe2018massively}) and contextual word embeddings pre-trained on masked language modeling, notably multilingual BERT (\cite{devlin2018bert}) and the cross-lingual language model (XLM) of \cite{lample2019cross}.  We use XLM as our baseline representation in all experiments, as it's the current state-of-the-art on the commonly used XNLI benchmark for cross-lingual understanding.

\subsection{Domain adaptation}
Domain adaptation, closely related to transfer learning, has a rich history in machine learning and natural language processing (\cite{pan2009survey}). Such methods have long been applied to document classification tasks \citep{blitzer2007biographies, glorot2011domain, al2017approaches, xu2017cross}.

Domain adaptation for NLP is intimately related to transfer learning and semi-supervised learning (\cite{chapelle2009semi}). Transfer learning has made tremendous advances recently due to the success of pre-training representations using language modeling as a source task (\cite{radford2018improving, peters2018deep, devlin2018bert}).  While such representations trained on large amounts of general domain text have been shown to transfer well generally, performance still suffers when the target domain is sufficiently different than what the models were pre-trained on.  In such cases, it is known that further pre-training the language model on in-domain text is helpful (\cite{howard2018universal, chronopoulou2019embarrassingly}).  It is natural to use unsupervised domain data for this task, when available (\cite{lee2019biobert, han2019unsupervised}).  

The study of weakly supervised learning in language processing is relatively new \citep{johnson2016supervised, yu2018diverse}.  Most recently, \cite{xie2019unsupervised} has introduced an unsupervised data augmentation (UDA) technique to demonstrate improvements in the few-shot learning setting.  Here, we extend this technique to facilitate cross-lingual and cross-domain transfer.

\section{Framework}
\label{sec:framework}

In this section, we formally define the problem discussed in this paper and describe the proposed approach in detail. 

\subsection{Problem Formulation}
\label{sec:formulation}

In vanilla zero-shot cross-lingual document classification, it is assumed that we have available a labeled dataset in a source language (English in our case), which we can denote by $L_{src} = \{(x,y) | x \sim P_{src}(x) \}$, where $P_{src}(x)$ is the prior distribution of task data in the source language.  It is assumed that no data is available for the task in the target language.  General purpose parallel and monolingual resources are used to train a cross-lingual classifier on the labeled source data, which is then applied to the target language data at test time.

In this work, we also assume access to a large unlabeled corpus in the target language, $U_{tgt} = \{x | x \sim P_{tgt}(x) \}$, which is usually the case in practical applications.  We aim to utilize this domain-specific unlabeled data to learn the best classifier for the data from the target language.  We refer to this setting as \textit{semi-supervised XLU}, although we're still in the zero-shot setting, in that no labeled data is used in the target language.

\subsection{Baseline Approaches}
There are two standard ways to transfer knowledge across the languages in the vanilla zero-shot setting: (1) using a translation system to translate the labeled samples, such as translate-train and translate-test methods, and (2) learning a multilingual embedding system to obtain a language irrelevant representations of the data.  
In this paper, we adopt the second approach as the basic model, and utilize the XLM model \citep{lample2019cross} as our base model, which has been pre-trained by large-scale parallel and monolingual data from various languages. 
Because XLM is a multilingual embedding system, a baseline is obtained by fine-tuning XLM with the labeled set $L_{src}(x)$ and directly applying the resulting model to the target language. In the experiments section, we also discuss the combination of the XLM and the translation based approaches.

\subsection{Semi-supervised XLU}
\label{sec:uda}

As argued in Introduction, even with a perfect translation or multilingual embedding system, we still face the domain-mismatch problem.     
This mismatch may limit the generalization ability of the model during testing time. 
To fully adapt the classifier to the target distribution, we explore the following approaches, each of which leverages unlabeled data in the target language in different ways.

\paragraph{Masked Language Model pre-training} BERT \citep{devlin2018bert} and its derivations (such as XLM) are trained on general domain corpora.  It is standard practice to further pre-train to adapt to a particular domain when data is available.  This technique can lead to improved performance for the target domain. We refer to this approach as the MLM pre-training.

However, in the cross-lingual setting, fine-tuning the XLM model in the target language can make the model degenerate in the source language, decreasing its ability to transfer across languages.  Therefore, in this case, we take care to use this method in combination with the translate-train method, which translates all labeled samples into the target language.

\paragraph{Unsupervised Data Augmentation} The second approach is utilizing the state-of-the-art semi-supervised learning technique, Unsupervised Data Augmentation (UDA) algorithm \citep{xie2019unsupervised}, to leverage the unlabeled data. The objective function of UDA can be written as,
\begin{equation}
    \min_{\theta} \E_{(x,y) \in L_{src}} \left[ -\text{log}(p_\theta(y | x)) \right] + \lambda \E_{x \in U_{tgt}} \left[ D_{KL} (p_\theta(y|\hat{x}) || p_\theta(y|x)) \right]
\label{eq:uda}
\end{equation}
where $\hat{x}$ is an augmented sample generated by a predefined augmentation function $\hat{x} = q(x)$. 
The augmentation function can be a paraphrase generation model, or a noising heuristic.  
Here, we use a machine translation system for this purpose.
The UDA loss enforces the classifier to produce label consistent predictions for pairs of original and augmented samples. 

In the cross-lingual setting, there are multiple ways of generating augmented samples using translation.  One could translate samples from the target language into the source language and use this cross-lingual pair as the augmented sample.  Alternatively, one could translate back into the target language and use only target-language augmented samples.  We find that the latter works best.
It is also possible to do data augmentation using source domain unlabeled data.  The results of these comparisons are included in out detailed ablation study in the experiments section.

\begin{figure}
    \centering
    \includegraphics[width=\linewidth]{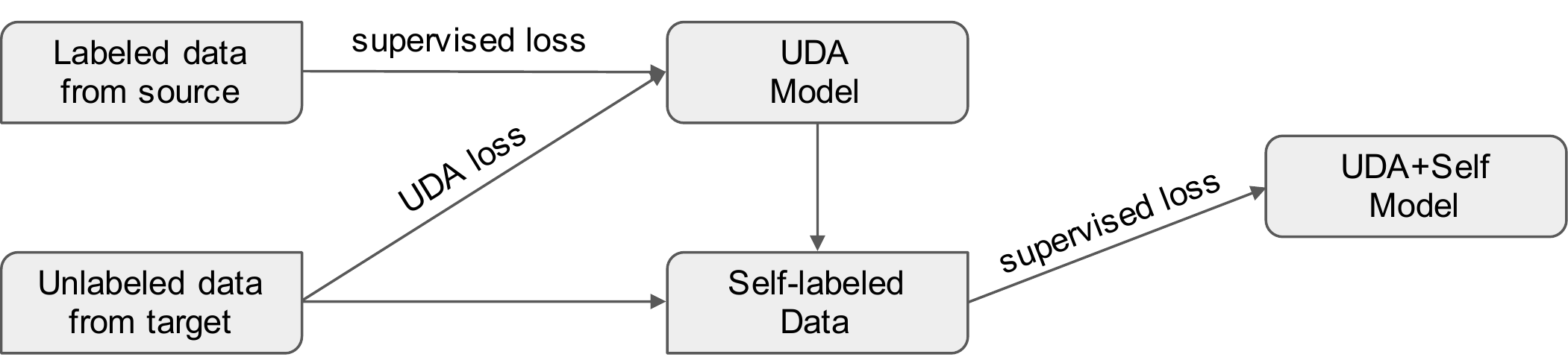}
    \caption{The process diagram of the proposed framework.}
    \label{fig:process}
\end{figure}

\paragraph{Alleviating the Train-Test Discrepancy of the UDA Method}
With the UDA algorithm, the classifier is able to learn some prior information on the target domain, however it still suffers from the train-test discrepancy. 
During the testing phase, our goal is to maximize the classifier performance on the target language, which can be written as,
\begin{equation}
    \max_{\theta} \E_{x \sim P_{tgt}(x)} p_\theta(y|x)
\label{eq:test}
\end{equation}
where $y$ is the ground-truth label of the sample $x$. 
Upon observing the training objective of the UDA method, Eq. \eqref{eq:uda}, one can see that the data $x$ that feed to model in the training phrase is sampled from three domains: (1) the source domain $P_{src}(x)$, (2) the target domain $P_{tgt}(x)$ and (3) the augmented sample domain $P_{aug}(x)$. 
On the other hand, the testing phrase only processes data from the target domain $P_{tgt}(x)$.

The source and target domain are mismatched, due to differences in language as argued earlier.  Furthermore, the augmented domain, although generated from the target domain, can also be mismatched, due to artifacts introduced by the translation system.  This can be especially problematic, since the UDA method needs diversity in the augmented samples to perform well \citep{xie2019unsupervised}, which trades off against their quality.

We propose to apply the self-training technique \citep{lee2013pseudo} to tackle this problem. 
We first train a classifier based on the UDA algorithm and denote it as $f^*(x)$, which is the teacher model used to score the unlabeled data $U_{tgt}$ from the target domain.  
Then we fine-tune a new XLM model using the soft classification loss function with the pseudo-labeled data, which is written as,
\begin{equation}
    \min_\theta \E_{x \in U_{tgt}} D_{KL} (f_\theta(x) || f^*(x))
\label{eq:self}
\end{equation}
Follwing this process, we obtain a new classifier trained only based on the target domain, which does not suffer from the train-test mismatch problem. We show that this process provides better generalization ability compared to the teacher model. A process diagram of the final model is presented in figure \ref{fig:process}.

\section{Experiments}
\label{sec:experiment}

In this section, we present a comprehensive study on two benchmark tasks, cross-lingual sentiment classification, and cross-lingual news classification.

\subsection{Datasets}

\paragraph{Sentiment Classification}
In this task, we test the proposed framework on a sentiment classification benchmark in three target languages, i.e. French, German and Chinese. 
The French, German and English data come from the benchmark cross-lingual Amazon reviews dataset \citep{prettenhofer2010cross}, which we denote as \textit{amazon-fr}, \textit{amazon-de} and \textit{amazon-en}. 
We merge training and testing samples from all product categories in one language, which leads to 6000 training samples. 
However, for the purpose of facilitating fair comparison with previous work, we also provide results for specific categories. 
In addition, we use 54K unlabeled samples from amazon-fr and 310K unlabeled samples from amazon-de.

For Chinese, we use the Chinese Amazon (\textit{amazon-cn}) \citep{zhang2015daily} and Dianping \citep{zhang2014explicit} datasets. Dianping is a business review website similar to Yelp. The training data for amazon-cn is amazon-en, and for dianping it is the Yelp dataset \citep{zhang2015character}. 
In these two cases, the size of the training sample is 2000. For both amazon-cn and dianping datasets, we have 4M unlabeled examples. 
Because the number of the unlabeled set is very large, we randomly sample 10\% for the UDA algorithm. 

\paragraph{News Classification} We use the MLDoc dataset \citep{schwenk2018corpus} for this task. The MLdoc dataset is a subset of RCV2 multilingual news dataset \citep{lewis2004rcv1}. It has 4 categories, i.e.  Corporate/Industrial, Economics, Government/Social and Markets, and each category has 250 training samples. We use the rest of the news documents in RCV2 dataset as the unlabeled data. The number of unlabeled samples for each language ranges from 5K to 20K, which is relatively smaller compared to the sentiment classification task. 

Because the XLM model is pre-trained on 15 languages, we ignore languages which are not supported by XLM in the above benchmark datasets. 

The pre-processing scripts for the above datasets, augmented samples and experiment settings needed to reproduce results are released in the Github repo \footnote{https://github.com/laiguokun/xlu-data}. 

\subsection{Masked language model pre-training strategy}

As introduced in section \ref{sec:uda}, we apply MLM pre-training on the unlabeled data corpus to obtain a domain-specific XLM, denoted as XLM$_{ft}$ in the following sections. The pre-training strategies for the two tasks are slightly different. 
In the sentiment classification task, because the size of the unlabeled corpus in each target domain is large enough, we fine-tune an XLM with MLM loss for each target domain respectively. 
In contrast, we do not have enough unlabeled data in each language in the MLDoc dataset, therefore we integrate unlabeled data from all languages as the training corpus.
As a result the XLM$_{ft}$ still preserves its language universality in this task.

\subsection{Main Results}
\label{sec:results}
We compare the follwing models:

\begin{itemize}[leftmargin=*]
    \item Fine-tune (Ft): Fine-tuning the pre-trained model with the source-domain training set.  In the case of XLM$_{ft}$, the training set is translated into the target language.
    \item Fine-tune with UDA (UDA): This method utilizes the unlabeled data from the target domain by optimizing the UDA loss function (Eq. \eqref{eq:uda}).
    \item Self-training based on the UDA model (UDA+Self): We first train the Ft model and UDA model, and choose the better one as the teacher model. The teacher model is used to train a new XLM student using only unlabeled data $U_{tgt}$ in the target domain, as described above.
\end{itemize}

We report the results of applying these three methods on both the original XLM model and the XLM$_{ft}$ model. In order to keep the notation simple, we use parenthesis after the method name to indicate which basic model was used, such as UDA(XLM$_{ft}$).  The details about the implementation and hyper-parameter tuning are included in Appendix \ref{app:details}.

The results for the cross-lingual sentiment classification task are summarized in table \ref{tab:sentiment1}. 
As our experiment setting on the cross-lingual amazon dataset is different from previous publications,
in order to provide a fair comparison with previous works, we summarize the results of the standard category-wise setting in table \ref{tab:sentiment2}. 
The results for cross-lingual news classification is included in table \ref{tab:news}. The last column ``Unlabeled'' in these tables indicates whether this method utilizes the unlabeled data. 
For the monolingual baselines, the models are trained with labeled data from the target domain. The size of the labeled set is the same as the English training set used for cross-lingual experiments. 

We can summarize our findings as follows: 
\begin{itemize}[leftmargin=*]
\item Looking at Ft(XLM) results, it is clear that without the help of unlabeled data from the target domain, there still exists a substantial gap between the model performance of the cross-lingual settings and the monolingual baselines, even when using state-of-the-art pre-trained cross-lingual representations.  
\item  Both the UDA algorithm and MLM pre-training can offer significant improvements by utilizing the unlabeled data. 
In the sentiment classification task, where the unlabeled data size is larger, Ft(XLM$_{ft}$) model usnig MLM pre-training consistently provides larger improvements compared with the UDA method. 
On the other hand, the MLM method is relatively more resource intensive and takes longer to converge (see Appendix \ref{sec:time}). In contrast, in the MLdoc dataset, when the size of the unlabeled samples is limited, the UDA method is more helpful. 
\item The combination of both methods - as in the UDA(XLM$_{ft}$) model - consistently outperforms either method alone. In this case the additional improvement provided by the UDA algorithm is smaller, but still consistent.
\item In the sentiment classification task, we observe the self-training technique consistently improves over its teacher model. 
It offers best results in both XLM and XLM$_{ft}$ based classifiers. 
The results demonstrate that self-training process is able to alleviate the train-test distribution mismatch problem and provide better generalization ability.
In the MLdoc dataset, self-training also achieves the best results overall, however the gains are less clear.  We hypothesize that this technique is not as useful without enough number of unlabeled samples. 
\item Finally, comparing with the best cross-lingual results and monolingual fine-tune baseline, we are able to completely close the performance gap by utilizing unlabeled data in the target language. 
Furthermore, our framework reaches new state-of-the-art results, improving over vanilla XLM baselines by 44\% on average. 
\end{itemize}

Furthermore, we provide an additional baseline, which only uses English samples to perform semi-supervised learning, whose details are in Appendix \ref{app:english-baseline}. The experment results show that it lags behind the ones using unlabeled data from the target domain. This observation also justifies the importance of information from the target domain in the XLU task. 

\begin{table}[!ht]
\centering
    \begin{tabular}{l|c|cccc|c}
        \toprule
        Base Model & Train    & amazon-en & amazon-en & amazon-en & yelp     & Unlabeled \\
        & Test     & amazon-fr & amazon-de & amazon-cn & dianping &\\
        \midrule
        & Ft       & 11.35     & 12.75     & 15.5     &  12.66    & \xmark\\
        XLM        & UDA      & 9.83      & 10.08     & 11.92      &  8.37   & \cmark\\
        & UDA+Self & 9.03      & 9.32      & 11.19      &  7.8   & \cmark\\
        \midrule
        & Ft       & 7.67      & 6.7       & 11.32      &  6.15   & \cmark\\
        XLM$_{ft}$     & UDA      & 7.15      & 6.3       & 11.92      &  5.21   & \cmark\\
        & UDA+Self & \textbf{6.67}      & \textbf{5.77}      & \textbf{9.46}       &  \textbf{4.8}    & \cmark\\
        \midrule
        \midrule
        \multicolumn{7}{c}{Monolingual Baselines}  \\
        \midrule
        XLM & Ft & 8.32 & 10.67 & 11.69 & 8.73 & \xmark \\
        \midrule 
        XLM$_{ft}$ & UDA &  5.95 & 6.12 & 7.74 & 4.64 & \cmark \\ 
        \bottomrule
    \end{tabular}
    \caption{Error rates for the sentiment classification task.}
\label{tab:sentiment1}
\end{table}

\begin{table}[!ht]
\centering
    \begin{tabular}{l|c|ccccc|c}
        \toprule
        Base Model & Train         &       &       & en    &       &       & Unlabeled \\
        & Test          & fr    & de    & es    & zh    & ru    &           \\
        \midrule
        & pre-XLM sota & 21.97 & 13.75 & 20.7  & 25.27 & 33.22 & \xmark         \\
        \midrule
        & Ft            & 7.95  & 6.03  & 12.08 & 11.95 & 26.95 & \xmark          \\
        XLM        & UDA           & 4.97  & 4.3   & 8.67  & 9.85  & 27.22 & \cmark          \\
        & UDA+Self      & 4.75  & 4.35  &  8.78     & 9.7   & 27.68 & \cmark          \\
        \midrule
        & Ft            & 6.3   & 6.43  & 8.7   & 11.02 & 25.78 & \cmark          \\
        XLM$_{ft}$     & UDA           & 4.65  & 4.63  & 6.97  & \textbf{7.15}  & 16.5  & \cmark          \\
        & UDA+Self      & \textbf{4.6}   & \textbf{4.27}  & \textbf{6.53}      & \textbf{7.15}  & \textbf{15.65} & \cmark  \\
        \midrule
        \midrule
        \multicolumn{8}{c}{Monolingual Baselines}  \\
        \midrule
        XLM & Ft & 5.35 & 3.77 & 3.95 & 7.85 & 11.05 &  \xmark \\
        \midrule 
        XLM$_{ft}$ & UDA &  3.95 & 3.05 & 3.2 & 6.68 & 10.3 &\cmark \\ 
        \bottomrule    
    \end{tabular}
    \caption{Error rates for news document classification. The pre-XLM sota results are provided by \cite{artetxe2018massively}}
\label{tab:news}
\end{table}

\begin{table}[!ht]
\centering
    \begin{tabular}{l|c|ccc|ccc|c}
        \toprule
        Base Model & dataset         & \multicolumn{3}{|c|}{amazon-fr} & \multicolumn{3}{|c|}{amazon-de}  & Unlabeled         \\
        & category        & book & dvd    & music & book & dvd    & music & \\
        \midrule
        & pre-XLM sota   & 14      & 14.3      & 14       & 13.6    & 13.9      & 12.2 &  \cmark  \\
        \midrule
        XLM & UDA+Self   & 7.75    & 8.7       & 9.35     & 8.3     & 10        & 9.96 &  \cmark  \\
        XLM$_{ft}$ & UDA+Self & \textbf{6.5}     & \textbf{7.05}      & \textbf{7.15}     & \textbf{4.8}     & \textbf{7.1}       & \textbf{5.55} & \cmark \\    
        \bottomrule
    \end{tabular}
    \caption{Error rates for the sentiment classification task by product category. The pre-XLM sota results are provided by \cite{chen2019emoji}.}
\label{tab:sentiment2}
\end{table}

\subsection{Labeled Data in the Source Language has Limited Value}

\begin{figure}
    \centering
    \includegraphics[width=0.4\linewidth]{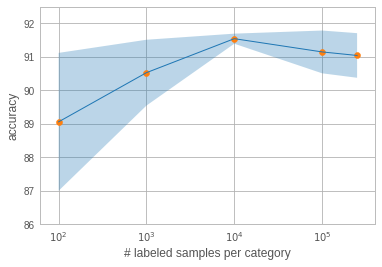}
    \includegraphics[width=0.4\linewidth]{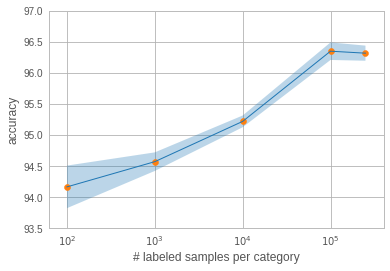}
    \caption{Plot of the accuracy vs. the number of training samples in log scale when running UDA(XLM) method. On the left is the cross-lingual setting. On the right is the monolingual setting.}
    \label{fig:num_tr}
\end{figure}

In this section, we provide evidence for the train-test domain discrepancy in the context of the UDA method, by showing that adding more labeled data in the source language does not improve target task accuracy after a certain point.
Figure \ref{fig:num_tr} plots the target model performance vs. the number of labeled training samples in the cross-lingual and monolingual settings respectively. 
The figures are based on the UDA(XLM) method with 6 runs in the Yelp-Dianping cross-lingual setting. 
The dot is the average accuracy and the filling area contains one standard derivation. 

We observe that, in the cross-lingual setting, the model performance peaks at around 10k training samples per category, and becomes worse with the larger training set. 
In contrast, the performance of the model improves consistently with more labeled data in the monolingual setting.  
This suggests that more training data from the source domain could harm model generalization ability in the target domain with UDA approach in the cross-lingual setting. 
In order to alleviate this issue, we propose to utilize the self-training technique, which abandons the data from the source domain and the augmentation domain, to maximize its performance in the target domain.

\subsection{Ablation study: Augmentation Strategies}

Next, we explore different augmentation strategies and their influence on the final performance. 
As stated in section \ref{sec:uda}, the augmentation strategy used in the main experiment is that we first translate the samples into English and translate them back to its original language.  We refer to this strategy as augmenting "from target domain to target domain" and abbreviate it as \textbf{t2t}.
 
We also explore two additional augmentation strategies:
(1) First, we do not translate the samples back to the target language and directly use English samples as the augmented samples, denoted as \textbf{t2s}.  
Naturally, the parallel samples in two languages have the same sentiment information and different input format which are suitable to be used as the augmentation sample pairs for the multilingual system such as XLM.
(2) The second approach is to leverage unlabeled data from other language domains. 
Here, we attempt to use the English unlabeled data. We translate them into the target language as the augmented samples. This strategy is denoted as \textbf{s2t}.

\begin{table}[!ht]
\centering
    \begin{tabular}{l|ccc||ccccc}
    \toprule
    train  &          & amazon-en &          &       \multicolumn{5}{c}{MLDoc en}          \\
    test   & amazon-fr & amazon-de & amazon-cn & fr   & de   & es   & zh   & ru    \\
    \midrule
    \textbf{t2t}       & 9.83     & \textbf{10.08}    & 11.92    & \textbf{4.65} & 4.63 & \textbf{6.97} & \textbf{8.15} & \textbf{16.5}  \\
    \textbf{t2s}       & 10.97    & 13.15    & 15.07    & 5.03 & 5    & 9.38 & 8.35 & 27.9  \\
    \textbf{t2t} + \textbf{t2s} & 9.82     & 12.07    & 11.93    & 5.07 & \textbf{4.33} & 7.15 & 8.9  & 28.6  \\
    \midrule
    \textbf{s2t}       & 9.32     & 10.38    & 12.28    & 5.53 & 6.25 & 7.73 & 8.5  & 29.17 \\
    \textbf{t2t} + \textbf{s2t} & \textbf{8.98}     & 10.87    & \textbf{11.79}    & 4.85 & 4.4  & 7.75 & 8.27 & 30.33 \\
    \bottomrule
    \end{tabular}
    \caption{Error rates when using different augmentation strategies and their combinations. Results for sentiment classification shown on the left, and news document classification on the right.}
    \label{tab:aug}
\end{table}

Table \ref{tab:aug} summarizes the performance of the proposed augmentation strategies and their combinations with the UDA(XLM) method in the sentiment classification and the UDA(XLM$_{ft}$) in the news classification settings.
From the results, we conclude that \textbf{t2t} is the best performing approach, as it's the best matched to the target domain. Leveraging the unlabeled data from other domains does not offer consistent improvement, however can provide additional value in isolated cases. 

We include additional ablations regarding translation system in the appendix, including the application of translate-train method in our experiments (section \ref{app:ablationtranslate}) and effects of hyper-parameters (section \ref{app:temp}). 

\section{Conclusion}

In this paper, we tackled the domain mismatch challenge in cross-lingual document classification - an important, yet often overlooked problem in cross-lingual understanding.  We provided evidence for the existence and importance of this problem, even when utilizing strong pre-trained cross-lingual representations.

We proposed a framework combining cross-lingual transfer techniques with three domain adaptation methods; unsupervised data augmentation, masked language model pre-training and self-training, which can leverage unlabeled data in the target language to moderate the domain gap.  Our results show that by removing the domain discrepancy, we can close the performance gap between cross-lingual transfer and monolingual baselines almost completely for the document classification task.  We are also able to improve the state-of-the-art in this area by a large margin.  

While document classification is by no means the most challenging task for XLU, we believe the strong gains that we demonstrated will encourage the community to pay more attention to domain drift in the cross-lingual setting.  Developing cross-lingual methods which are competitive with in-language models for real world, semantically challenging NLP problems remains an open problem and subject of future research.
\bibliography{iclr2020_conference}

\begin{thebibliography}{37}
\providecommand{\natexlab}[1]{#1}
\providecommand{\url}[1]{\texttt{#1}}
\expandafter\ifx\csname urlstyle\endcsname\relax
  \providecommand{\doi}[1]{doi: #1}\else
  \providecommand{\doi}{doi: \begingroup \urlstyle{rm}\Url}\fi

\bibitem[Al-Moslmi et~al.(2017)Al-Moslmi, Omar, Abdullah, and
  Albared]{al2017approaches}
Tareq Al-Moslmi, Nazlia Omar, Salwani Abdullah, and Mohammed Albared.
\newblock Approaches to cross-domain sentiment analysis: A systematic
  literature review.
\newblock \emph{IEEE Access}, 5:\penalty0 16173--16192, 2017.

\bibitem[Aly et~al.(2018)Aly, Lakhotia, Zhao, Mohit, Oguz, Arora, Gupta, Dewan,
  Nelson-Lindall, and Shah]{aly2018pytext}
Ahmed Aly, Kushal Lakhotia, Shicong Zhao, Mrinal Mohit, Barlas Oguz, Abhinav
  Arora, Sonal Gupta, Christopher Dewan, Stef Nelson-Lindall, and Rushin Shah.
\newblock Pytext: A seamless path from nlp research to production.
\newblock \emph{arXiv preprint arXiv:1812.08729}, 2018.

\bibitem[Artetxe \& Schwenk(2018)Artetxe and Schwenk]{artetxe2018massively}
Mikel Artetxe and Holger Schwenk.
\newblock Massively multilingual sentence embeddings for zero-shot
  cross-lingual transfer and beyond.
\newblock \emph{arXiv preprint arXiv:1812.10464}, 2018.

\bibitem[Bel et~al.(2003)Bel, Koster, and Villegas]{bel2003cross}
Nuria Bel, Cornelis~HA Koster, and Marta Villegas.
\newblock Cross-lingual text categorization.
\newblock In \emph{International Conference on Theory and Practice of Digital
  Libraries}, pp.\  126--139. Springer, 2003.

\bibitem[Blitzer et~al.(2007)Blitzer, Dredze, and
  Pereira]{blitzer2007biographies}
John Blitzer, Mark Dredze, and Fernando Pereira.
\newblock Biographies, bollywood, boom-boxes and blenders: Domain adaptation
  for sentiment classification.
\newblock In \emph{Proceedings of the 45th annual meeting of the association of
  computational linguistics}, pp.\  440--447, 2007.

\bibitem[Chapelle et~al.(2009)Chapelle, Scholkopf, and Zien]{chapelle2009semi}
Olivier Chapelle, Bernhard Scholkopf, and Alexander Zien.
\newblock Semi-supervised learning (chapelle, o. et al., eds.; 2006)[book
  reviews].
\newblock \emph{IEEE Transactions on Neural Networks}, 20\penalty0
  (3):\penalty0 542--542, 2009.

\bibitem[Chen et~al.(2019)Chen, Shen, Hu, Lu, Mei, and Liu]{chen2019emoji}
Zhenpeng Chen, Sheng Shen, Ziniu Hu, Xuan Lu, Qiaozhu Mei, and Xuanzhe Liu.
\newblock Emoji-powered representation learning for cross-lingual sentiment
  classification.
\newblock In \emph{The World Wide Web Conference}, pp.\  251--262. ACM, 2019.

\bibitem[Chronopoulou et~al.(2019)Chronopoulou, Baziotis, and
  Potamianos]{chronopoulou2019embarrassingly}
Alexandra Chronopoulou, Christos Baziotis, and Alexandros Potamianos.
\newblock An embarrassingly simple approach for transfer learning from
  pretrained language models.
\newblock \emph{arXiv preprint arXiv:1902.10547}, 2019.

\bibitem[Conneau et~al.(2018)Conneau, Lample, Rinott, Williams, Bowman,
  Schwenk, and Stoyanov]{conneau2018xnli}
Alexis Conneau, Guillaume Lample, Ruty Rinott, Adina Williams, Samuel~R Bowman,
  Holger Schwenk, and Veselin Stoyanov.
\newblock Xnli: Evaluating cross-lingual sentence representations.
\newblock \emph{arXiv preprint arXiv:1809.05053}, 2018.

\bibitem[Devlin et~al.(2018)Devlin, Chang, Lee, and Toutanova]{devlin2018bert}
Jacob Devlin, Ming-Wei Chang, Kenton Lee, and Kristina Toutanova.
\newblock Bert: Pre-training of deep bidirectional transformers for language
  understanding.
\newblock \emph{arXiv preprint arXiv:1810.04805}, 2018.

\bibitem[Glorot et~al.(2011)Glorot, Bordes, and Bengio]{glorot2011domain}
Xavier Glorot, Antoine Bordes, and Yoshua Bengio.
\newblock Domain adaptation for large-scale sentiment classification: A deep
  learning approach.
\newblock In \emph{Proceedings of the 28th international conference on machine
  learning (ICML-11)}, pp.\  513--520, 2011.

\bibitem[Han \& Eisenstein(2019)Han and Eisenstein]{han2019unsupervised}
Xiaochuang Han and Jacob Eisenstein.
\newblock Unsupervised domain adaptation of contextualized embeddings: A case
  study in early modern english.
\newblock \emph{arXiv preprint arXiv:1904.02817}, 2019.

\bibitem[Howard \& Ruder(2018)Howard and Ruder]{howard2018universal}
Jeremy Howard and Sebastian Ruder.
\newblock Universal language model fine-tuning for text classification.
\newblock \emph{arXiv preprint arXiv:1801.06146}, 2018.

\bibitem[Johnson \& Zhang(2016)Johnson and Zhang]{johnson2016supervised}
Rie Johnson and Tong Zhang.
\newblock Supervised and semi-supervised text categorization using lstm for
  region embeddings.
\newblock \emph{arXiv preprint arXiv:1602.02373}, 2016.

\bibitem[Kingma \& Ba(2014)Kingma and Ba]{kingma2014adam}
Diederik~P Kingma and Jimmy Ba.
\newblock Adam: A method for stochastic optimization.
\newblock \emph{arXiv preprint arXiv:1412.6980}, 2014.

\bibitem[Klementiev et~al.(2012)Klementiev, Titov, and
  Bhattarai]{klementiev2012inducing}
Alexandre Klementiev, Ivan Titov, and Binod Bhattarai.
\newblock Inducing crosslingual distributed representations of words.
\newblock In \emph{Proceedings of COLING 2012}, pp.\  1459--1474, 2012.

\bibitem[Lample \& Conneau(2019)Lample and Conneau]{lample2019cross}
Guillaume Lample and Alexis Conneau.
\newblock Cross-lingual language model pretraining.
\newblock \emph{arXiv preprint arXiv:1901.07291}, 2019.

\bibitem[Lee(2013)]{lee2013pseudo}
Dong-Hyun Lee.
\newblock Pseudo-label: The simple and efficient semi-supervised learning
  method for deep neural networks.
\newblock In \emph{Workshop on Challenges in Representation Learning, ICML},
  volume~3, pp.\ ~2, 2013.

\bibitem[Lee et~al.(2019)Lee, Yoon, Kim, Kim, Kim, So, and
  Kang]{lee2019biobert}
Jinhyuk Lee, Wonjin Yoon, Sungdong Kim, Donghyeon Kim, Sunkyu Kim, Chan~Ho So,
  and Jaewoo Kang.
\newblock Biobert: pre-trained biomedical language representation model for
  biomedical text mining.
\newblock \emph{arXiv preprint arXiv:1901.08746}, 2019.

\bibitem[Lewis et~al.(2004)Lewis, Yang, Rose, and Li]{lewis2004rcv1}
David~D Lewis, Yiming Yang, Tony~G Rose, and Fan Li.
\newblock Rcv1: A new benchmark collection for text categorization research.
\newblock \emph{Journal of machine learning research}, 5\penalty0
  (Apr):\penalty0 361--397, 2004.

\bibitem[Liu et~al.(2019)Liu, Lin, Liu, and Sun]{liu2019xqa}
Jiahua Liu, Yankai Lin, Zhiyuan Liu, and Maosong Sun.
\newblock Xqa: A cross-lingual open-domain question answering dataset.
\newblock In \emph{Proceedings of the 57th Conference of the Association for
  Computational Linguistics}, pp.\  2358--2368, 2019.

\bibitem[Mikolov et~al.(2013)Mikolov, Le, and Sutskever]{mikolov2013exploiting}
Tomas Mikolov, Quoc~V Le, and Ilya Sutskever.
\newblock Exploiting similarities among languages for machine translation.
\newblock \emph{arXiv preprint arXiv:1309.4168}, 2013.

\bibitem[Nivre et~al.(2016)Nivre, De~Marneffe, Ginter, Goldberg, Hajic,
  Manning, McDonald, Petrov, Pyysalo, Silveira, et~al.]{nivre2016universal}
Joakim Nivre, Marie-Catherine De~Marneffe, Filip Ginter, Yoav Goldberg, Jan
  Hajic, Christopher~D Manning, Ryan McDonald, Slav Petrov, Sampo Pyysalo,
  Natalia Silveira, et~al.
\newblock Universal dependencies v1: A multilingual treebank collection.
\newblock In \emph{Proceedings of the Tenth International Conference on
  Language Resources and Evaluation (LREC 2016)}, pp.\  1659--1666, 2016.

\bibitem[Pan \& Yang(2009)Pan and Yang]{pan2009survey}
Sinno~Jialin Pan and Qiang Yang.
\newblock A survey on transfer learning.
\newblock \emph{IEEE Transactions on knowledge and data engineering},
  22\penalty0 (10):\penalty0 1345--1359, 2009.

\bibitem[Paszke et~al.(2017)Paszke, Gross, Chintala, Chanan, Yang, DeVito, Lin,
  Desmaison, Antiga, and Lerer]{paszke2017automatic}
Adam Paszke, Sam Gross, Soumith Chintala, Gregory Chanan, Edward Yang, Zachary
  DeVito, Zeming Lin, Alban Desmaison, Luca Antiga, and Adam Lerer.
\newblock Automatic differentiation in pytorch.
\newblock In \emph{NIPS-W}, 2017.

\bibitem[Peters et~al.(2018)Peters, Neumann, Iyyer, Gardner, Clark, Lee, and
  Zettlemoyer]{peters2018deep}
Matthew~E Peters, Mark Neumann, Mohit Iyyer, Matt Gardner, Christopher Clark,
  Kenton Lee, and Luke Zettlemoyer.
\newblock Deep contextualized word representations.
\newblock \emph{arXiv preprint arXiv:1802.05365}, 2018.

\bibitem[Prettenhofer \& Stein(2010)Prettenhofer and
  Stein]{prettenhofer2010cross}
Peter Prettenhofer and Benno Stein.
\newblock Cross-language text classification using structural correspondence
  learning.
\newblock In \emph{Proceedings of the 48th annual meeting of the association
  for computational linguistics}, pp.\  1118--1127, 2010.

\bibitem[Radford et~al.(2018)Radford, Narasimhan, Salimans, and
  Sutskever]{radford2018improving}
Alec Radford, Karthik Narasimhan, Tim Salimans, and Ilya Sutskever.
\newblock Improving language understanding by generative pre-training.
\newblock \emph{URL https://s3-us-west-2. amazonaws.
  com/openai-assets/researchcovers/languageunsupervised/language understanding
  paper. pdf}, 2018.

\bibitem[Ruder et~al.(2017)Ruder, Vuli{\'c}, and S{\o}gaard]{ruder2017survey}
Sebastian Ruder, Ivan Vuli{\'c}, and Anders S{\o}gaard.
\newblock A survey of cross-lingual word embedding models.
\newblock \emph{arXiv preprint arXiv:1706.04902}, 2017.

\bibitem[Schwenk \& Douze(2017)Schwenk and Douze]{schwenk2017learning}
Holger Schwenk and Matthijs Douze.
\newblock Learning joint multilingual sentence representations with neural
  machine translation.
\newblock \emph{arXiv preprint arXiv:1704.04154}, 2017.

\bibitem[Schwenk \& Li(2018)Schwenk and Li]{schwenk2018corpus}
Holger Schwenk and Xian Li.
\newblock A corpus for multilingual document classification in eight languages.
\newblock \emph{arXiv preprint arXiv:1805.09821}, 2018.

\bibitem[Xie et~al.(2019)Xie, Dai, Hovy, Luong, and Le]{xie2019unsupervised}
Qizhe Xie, Zihang Dai, Eduard Hovy, Minh-Thang Luong, and Quoc~V Le.
\newblock Unsupervised data augmentation.
\newblock \emph{arXiv preprint arXiv:1904.12848}, 2019.

\bibitem[Xu \& Yang(2017)Xu and Yang]{xu2017cross}
Ruochen Xu and Yiming Yang.
\newblock Cross-lingual distillation for text classification.
\newblock \emph{arXiv preprint arXiv:1705.02073}, 2017.

\bibitem[Yu et~al.(2018)Yu, Guo, Yi, Chang, Potdar, Cheng, Tesauro, Wang, and
  Zhou]{yu2018diverse}
Mo~Yu, Xiaoxiao Guo, Jinfeng Yi, Shiyu Chang, Saloni Potdar, Yu~Cheng, Gerald
  Tesauro, Haoyu Wang, and Bowen Zhou.
\newblock Diverse few-shot text classification with multiple metrics.
\newblock \emph{arXiv preprint arXiv:1805.07513}, 2018.

\bibitem[Zhang et~al.(2015{\natexlab{a}})Zhang, Zhao, and
  LeCun]{zhang2015character}
Xiang Zhang, Junbo Zhao, and Yann LeCun.
\newblock Character-level convolutional networks for text classification.
\newblock In \emph{Advances in neural information processing systems}, pp.\
  649--657, 2015{\natexlab{a}}.

\bibitem[Zhang et~al.(2014)Zhang, Lai, Zhang, Zhang, Liu, and
  Ma]{zhang2014explicit}
Yongfeng Zhang, Guokun Lai, Min Zhang, Yi~Zhang, Yiqun Liu, and Shaoping Ma.
\newblock Explicit factor models for explainable recommendation based on
  phrase-level sentiment analysis.
\newblock In \emph{Proceedings of the 37th international ACM SIGIR conference
  on Research \& development in information retrieval}, pp.\  83--92. ACM,
  2014.

\bibitem[Zhang et~al.(2015{\natexlab{b}})Zhang, Zhang, Zhang, Lai, Liu, Zhang,
  and Ma]{zhang2015daily}
Yongfeng Zhang, Min Zhang, Yi~Zhang, Guokun Lai, Yiqun Liu, Honghui Zhang, and
  Shaoping Ma.
\newblock Daily-aware personalized recommendation based on feature-level time
  series analysis.
\newblock In \emph{Proceedings of the 24th international conference on world
  wide web}, pp.\  1373--1383. International World Wide Web Conferences
  Steering Committee, 2015{\natexlab{b}}.

\end{thebibliography}


\begin{thebibliography}{3}
\providecommand{\natexlab}[1]{#1}
\providecommand{\url}[1]{\texttt{#1}}
\expandafter\ifx\csname urlstyle\endcsname\relax
  \providecommand{\doi}[1]{doi: #1}\else
  \providecommand{\doi}{doi: \begingroup \urlstyle{rm}\Url}\fi

\bibitem[Bengio \& LeCun(2007)Bengio and LeCun]{Bengio+chapter2007}
Yoshua Bengio and Yann LeCun.
\newblock Scaling learning algorithms towards {AI}.
\newblock In \emph{Large Scale Kernel Machines}. MIT Press, 2007.

\bibitem[Goodfellow et~al.(2016)Goodfellow, Bengio, Courville, and
  Bengio]{goodfellow2016deep}
Ian Goodfellow, Yoshua Bengio, Aaron Courville, and Yoshua Bengio.
\newblock \emph{Deep learning}, volume~1.
\newblock MIT Press, 2016.

\bibitem[Hinton et~al.(2006)Hinton, Osindero, and Teh]{Hinton06}
Geoffrey~E. Hinton, Simon Osindero, and Yee~Whye Teh.
\newblock A fast learning algorithm for deep belief nets.
\newblock \emph{Neural Computation}, 18:\penalty0 1527--1554, 2006.

\end{thebibliography}
\bibliographystyle{iclr2020_conference}
\appendix

\section{Appendix of Cross-ligual Experiments}

\subsection{Implementation Details}
\label{app:details}

The experiments in this paper are based on the PyTorch \citep{paszke2017automatic} and Pytext \citep{aly2018pytext} package. We use the Adam \citep{kingma2014adam} as the optimizer. 
For all experiments, we grid search the learning rate in the set $\{5 \times 10^{-6}, 1\times 10^{-5}, 2 \times 10^{-5}\}$. 
When using UDA method, we also try the three different annealing strategies introduced in the UDA paper \citep{xie2019unsupervised}, and the $\lambda$ in \eqref{eq:uda} is always set as 1. 
The batch size in the Ft and UDA+Self method is 128. In the UDA method, the batch size is 16 for the labeled data and 80 for the unlabeled data. 
Due to the limitation of the GPU memory, in all experiments, we set the length of samples as 256, and cut the input tokens exceeding this length. 
Finally, we report the results with the best hyper-parameters.

As for the augmentation process, we sweep the temperature which controls the diversity of beam search in translation. 
The best temperature for ``en-de, en-fr, en-es'' and  ``en-ru'' are 1.0 and 0.6, the sampling space is the whole vocabulary. 
In the ``en-zh'' setting, the temperature is 1.0 and the sampling space is the top 100 tokens in the vocabulary. 
We note that this uses the Facebook production translation models, and results could vary when other translation systems are applied.  For reproducibility, we will release the augmented datasets that we generated.

\subsection{Ablation Study: The Baseline with English Unlabeled Data}
\label{app:english-baseline}

Here, we provide a baseline only using English samples to perform semi-supervised learning. More specifically, we first train the model with English unlabeled data and augmented samples, then tests it on different target domains. 
This approach is similar to the traditional translate-test method.
This method offers a baseline, which merely increases the size of data but without providing the target domain information. 
During the test phrasing, we experiment with two input strategies. One is using the original test samples, and another is translating the samples into English. 
We report the results (Table \ref{tab:trtest}) of the UDA(XLM) method with two input strategies and compare them with the main results, which uses the unlabeled data from the target domain. 
First, we observe that the performance of using original and translated samples is similar. 
Second, compared with Ft(XLM) baselines in section \ref{sec:results}, utilizing the unlabeled data from the English domain is slightly better than only training with labeled data, but it still lags behind the performance of using the unlabeled data from the target domain.  

\begin{table}[!ht]
\centering
    \resizebox{1.\textwidth}{!}{
    \begin{tabular}{lc|ccc||ccccc}
    \toprule
    lang of test  & Unlabeled & amazon-fr & amazon-de & amazon-cn & fr   & de   & es   & zh   & ru    \\
    \midrule
    original & English & 10.57    & 12.98    & 13.73    & 6.47 & 6.13 & 12.35 & 11.6  & 27.52 \\
    translated & English & 10.68    & 12.4     & 15.21    & 6.2  & 6.2  & 11.88 & 10.12 & 31.37 \\
    \midrule
    \midrule
    original & Target & \textbf{9.83} & \textbf{10.08} & \textbf{11.92} & \textbf{4.97} & \textbf{4.3} & \textbf{8.67} & \textbf{9.85} & \textbf{27.22} \\
    \bottomrule
    \end{tabular}
    }
    \caption{The first part is the baseline results using the English unlabeled data. The second part is the results using the unlabeled data from the target domain, which are copied from the section \ref{sec:results}. }
    \label{tab:trtest}
\end{table}

\subsection{Ablation study: Translate-Train}
\label{app:ablationtranslate}

As discussed earlier, fine-tuning XLM on the target language would depreciate the multilingual ability of the model. We apply the translate-train method to tackle this problem. In order to understand the influence of this strategy when using the proposed framework, we perform an ablation study. We test 3 input strategies: (1) English: use the original English data as training data. (2) tr-train: use the translate-train strategy, which translate the training data into the target language. (3) both: we combine the (1) and (2) as the training data. We report the results of the UDA(XLM) method in the sentiment classification tasks and UDA(XLM$_{ft}$) method in the news classification tasks in table \ref{tab:trt-lang}.

\begin{table}[!ht]
\centering
    \begin{tabular}{l|ccc||ccccc}
    \toprule
    train &           & amazon-en      &           &      \multicolumn{5}{c}{MLDoc en}       \\
    test  & amazon-fr & amazon-de & amazon-cn & fr   & de   & es   & zh   & ru    \\
    \midrule
    English   & 9.83      & \textbf{10.08}     & \textbf{12.02}     & \textbf{4.75} & \textbf{4.63} & 8.07 & \textbf{8.15} & 28.28 \\
    tr-train   & 9.92      & 11.7      & 15.31     & 5.57 & 5    & 6.97 & 9.2  & \textbf{16.5}  \\
    both  & \textbf{9.75}      & 11.02     & 16.02     & 4.95 & 4.87 & \textbf{6.33} & 8.35 & 28.63 \\
    \bottomrule
    \end{tabular}
    \caption{Ablation Study about the translate-train strategies. Results for sentiment classification shown on the left, and news document classification on the right.}
    \label{tab:trt-lang}
\end{table}

We observe that in most cases, using the original training examples achieves the best performance. However, in special cases such as MLDoc-ru, the translate-train method achieves better performance. We recommend trying both approaches in practice.

\subsection{Ablation study: diversity in beam decoding of data augmentation}
\label{app:temp}

Given a translation system, we use the sample decoding strategy to translate the sample. 
The sample space is the entire vocabulary space. We tune the temperature of $\mu$ of the softmax distribution. 
As discussed in \cite{xie2019unsupervised}, this controls the trade-off between quality and diversity. 
When $\mu = 0$, the sampling reduces to the greedy search and produce the best quality samples.
When $\mu = 1$, the sampling produces diverse outputs but loses some semantic information. 
The table \ref{tab:temperature} illustrates the influence of $\mu$ value to the final performance in the English-to-French and English-to-German settings. 
The results show that temperature has a significant influence on the final performance. 
However, because the quality of translation systems for different language pairs are not the same, their best temperature also varies. 
In the appendix \ref{app:details}, we include the best temperature values for the translation systems used in this paper.  

\begin{table}[!ht]
\centering
    \begin{tabular}{l|cccc}
    \toprule
    $\mu$ & amazon-fr & amazon-de & mldoc-fr & mldoc-de \\
    \midrule
    0.6  & 11.92    & 13.27   & 8.95    & 7.4     \\
    0.8  & \textbf{9.63}     & 10.6    & 5.55    & 4.7     \\
    1.0    & 9.83     & \textbf{10.08}   & \textbf{4.97}    & \textbf{4.3}     \\
    1.2  & 10.83    & 13.58   & 7.87    & 5.5    \\
    \bottomrule
    \end{tabular}
    \label{tab:temperature}
    \caption{Effect of the temperature of the translation sampling decoder.}
\end{table}

\subsection{Computation Time of UDA and MLM Pretraining}
\label{sec:time}

From the main results in section \ref{sec:results}, we can see that MLM pre-training can offer better improvements over the UDA method. However, it is also more resource intensive, since MLM pre-training is a token level task with a large output space, which leads to more computationally intensive updates and also takes longer to converge.
In our experiments, we used NVIDIA V100-16G GPUs to train all models. 8 GPUs were used to train Ft and UDA methods, and 32 GPUs to perform MLM pretraining. In the "amazonen->amazonfr" setting, for example, the unlabeled set contains 50K unlabeled samples and 8M tokens after BPE tokenization. The Ft method takes 3.2 GPU hours to converge. The UDA method training takes 16.8 GPU hours, excluding the time it takes to generate augmented samples (which we handle as part of data pre-processing). MLM pre-training takes upwards of 500 GPU hours to converge.  This is another factor which should be taken into account.

\section{Monolingual domain adaptation}

As further evidence that our method addresses the domain mismatch, we apply out framework to the monolingual cross-domain document classification problem. We again focus on sentiment classification where data comes from two different domains, product reviews (amazon-en, amazon-cn)  and business reviews (Yelp and Dianping). We train and test on the same language, only transferring across domains. We consider the two domain-pairs, amazonen-yelp and amazoncn-dianping. The results are illustrated in table \ref{tab:cross-domain}.  Conclusions are similar to the cross-domain setting (section \ref{sec:results}):
\begin{itemize}[leftmargin=*]
    \item There exists a clear gap between the cross-domain and in-domain results of the Ft method, even when using strong pre-trained representations. 
    \item By leveraging the unlabeled data from the target domain, we can significantly boost the model performance.
    \item Best results are achieved with our combined approach and almost completely matches the in-domain baselines.
\end{itemize}

\begin{table}[!ht]
\centering
    \begin{tabular}{l|c|cccc|c}
    \toprule
    Base Model & Train    & amazon-en & yelp     & amazon-cn & dianping  & Unlabeled \\
          & Test     & yelp     & amazon-en & dianping & amazon-cn  &\\
    \midrule
          & Ft       & 8.46     & 14.52    & 12.86    & 14.96    & \xmark \\
    XLM   & UDA      & 5.43     & 10.1     & 8.31     & 11.19    & \cmark \\
          & UDA+Self & 4.94     & 9.58     & 8.71     & 10.69    & \cmark \\
    \midrule
          & Ft       & 4.51     & 10.8     & 7.06     & 9.96     & \cmark \\
    XLM$_{ft}$ & UDA      & 3.78     & 7.87     & 5.54     & \textbf{7.38}     & \cmark \\
          & UDA+Self & \textbf{ 3.36}     & \textbf{7.47}     & \textbf{5.34}     & 7.4      & \cmark \\
    \midrule
    \midrule
    \multicolumn{7}{c}{In-Domain Baselines}  \\
    \midrule
    XLM & Ft & 5.8 & 10.9 & 11.43  & 12.31 & \xmark \\
    \midrule 
    XLM$_{ft}$ & UDA &  3.34 & 7.57 & 4.64 & 7.74 & \cmark \\ 
    \bottomrule
    \end{tabular}
    \caption{Error rates for cross-domain document classification}
    \label{tab:cross-domain}
\end{table}
\end{document}